\documentclass[a4paper,fleqn]{cas-dc}

\usepackage[authoryear]{natbib}
\usepackage{hyperref}
\usepackage{float}
\usepackage{verbatim} 
\usepackage{apalike}
\usepackage{xspace}
\usepackage{graphicx} 
\usepackage{hyperref} 
\usepackage{amsmath}
\usepackage{amssymb}
\usepackage{enumitem}
\usepackage{url}
\hypersetup{
    colorlinks=true,
    linkcolor=blue,
    filecolor=magenta,   
    citecolor=black, 
    urlcolor=black,
    }
\usepackage{comment}
\usepackage{multirow}
\usepackage{geometry}
\graphicspath{ {./figures/} }

\def\tsc#1{\csdef{#1}{\textsc{\lowercase{#1}}\xspace}}
\tsc{WGM}
\tsc{QE}
\tsc{EP}
\tsc{PMS}
\tsc{BEC}
\tsc{DE}

\begin{document}
\let\WriteBookmarks\relax
\def\floatpagepagefraction{1}
\def\textpagefraction{.001}
\shorttitle{NVP-HRI}
\shortauthors{Yuzhi Lai, et~al.}

\title [mode = title]{NVP-HRI: Zero Shot Natural Voice and Posture-based Human-Robot Interaction via Large Language Model}                      



\author[1]{Yuzhi Lai}[type=editor,
                        auid=001,
                        bioid=,
                        prefix=,
                        role=,
                        orcid=0009-0002-9976-7422]

\ead{Yuzhi.Lai@Reutlingen-University.DE}

\credit{Conceptualization, development of Methodology and Software}

\affiliation[1]{organization={University Reutlingen},
                addressline={Alteburgstraße 150}, 
                city={Reutlingen},
                postcode={72762}, 
                country={Germany}}

\author[2]{Shenghai Yuan}[type=editor,
                        auid=002,
                        bioid=,
                        prefix=,
                        role=,
                        orcid=0009-0003-1887-6342]
\cormark[1]
\ead{shyuan@ntu.edu.sg}

\credit{Conceptualization of this study, Writing - Original draft preparation}
\affiliation[2]{organization={Nanyang Technological University},
                addressline={50 Nanyang Avenue}, 
                city={Singapore},
                postcode={639798}, 
                country={Singapore}}

\author[1]{Youssef Nassar}[type=editor,
                        auid=003,
                        bioid=,
                        prefix=,
                        role=,
                        orcid=0009-0007-2867-4030]
\ead{Youssef.Nassar@Reutlingen-University.DE}
\credit{Data curation, Design and management of experiments, Validation, Writing - Original draft preparation}

\author[3]{Mingyu Fan}[type=,
                        auid=004,
                        bioid=,
                        prefix=,
                        role=,
                        orcid=0000-0002-0492-4708]
\ead{fanmingyu@dhu.edu.cn}
\credit{Data curation, Writing - Original draft preparation}

\affiliation[3]{organization={Donghua University},
                addressline={ 849 Zhongshan West Street 9}, 
                city={Shanghai},
                postcode={200051}, 
                country={China}}

\author[1]{Thomas Weber}[type=,
                        auid=005,
                        bioid=,
                        prefix=,
                        role=,
                        orcid=0000-0002-2932-2308]
\ead{Thomas.Weber@Reutlingen-University.DE}
\credit{Design and management of experiments, Writing - Original draft preparation}

\author[1]{Matthias Rätsch}[type=,
                        auid=007,
                        bioid=,
                        prefix=,
                        role=,
                        orcid=0000-0002-8254-8293]
\ead{Matthias.Raetsch@Reutlingen-University.DE}
\credit{Final approval of the version to be submitted}

\cortext[cor1]{Corresponding author}


\begin{abstract}
Effective Human-Robot Interaction (HRI) is crucial for future service robots in aging societies. Existing solutions are biased toward only well-trained objects, creating a gap when dealing with new objects. Currently, HRI systems using predefined gestures or language tokens for pretrained objects pose challenges for all individuals, especially elderly ones. These challenges include difficulties in recalling commands, memorizing hand gestures, and learning new names.
This paper introduces NVP-HRI, an intuitive multi-modal HRI paradigm that combines voice commands and deictic posture. NVP-HRI utilizes the Segment Anything Model (SAM) to analyze visual cues and depth data, enabling precise structural object representation. Through a pre-trained SAM network, NVP-HRI allows interaction with new objects via zero-shot prediction, even without prior knowledge.
NVP-HRI also integrates with a large language model (LLM) for multimodal commands, coordinating them with object selection and scene distribution in real time for collision-free trajectory solutions. We also regulate the action sequence with the essential control syntax to reduce LLM hallucination risks. The evaluation of diverse real-world tasks using a Universal Robot showcased up to 59.2\% efficiency improvement over traditional gesture control, as illustrated in the video \url{https://youtu.be/EbC7al2wiAc}. Our code and design will be openly available at \url{https://github.com/laiyuzhi/NVP-HRI.git}.
\end{abstract}




\maketitle

\section{Introduction}

\begin{figure*}[t]
      \centering
      \includegraphics[width=0.99\textwidth]{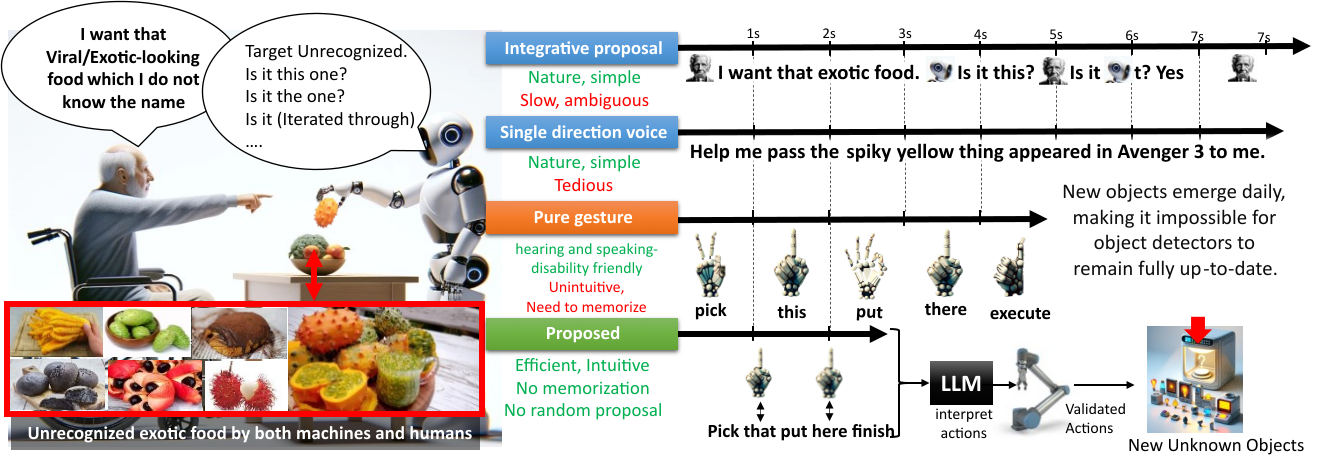}
      \caption{Proposed voice-posture fusion HRI method has superior efficiency in manipulating untrained objects and requires no memorization of key syntax, which is ideal for elderly and healthcare applications.}
      \label{motivationfigure}
   \end{figure*}

In a rapidly evolving society, the emergence of new objects poses challenges to individuals and robots, particularly in interacting with objects of unknown names. This is a significant hurdle for effective human-robot interaction (HRI), especially for the elderly or sick who rely on robot assistance. With an aging population and rising labor costs, there is growing interest in utilizing mobile manipulators \citep{li2023yolo, c2} or novel robot platforms \citep{gonccalves2023deep, cao2023doublebee, tang2024spatial} to address these challenges. A key factor for the success of future robotics services is ensuring that Human-Robot Interaction (HRI) remains intuitive and adaptable to a wide range of objects \citep{alonso2017identification, lu2023co,nguyen2024uloc}. However, existing HRI frameworks heavily rely on pre-trained detector networks \citep{esfahani2020unsupervised,GuLKC22,yang2024av,xu2024efficient} or often require mastering complex gestures or are prone to ambiguity when describing complex scenarios \citep{c4, hanggesture, gamboa2023asynchronous}, which is impractical for the elderly and even for ordinary people. There are significant domain gaps to bridge for intuitive HRI with rare objects \citep{yang2022overcoming,cao2023multi1,cao2023mopa,cao2024reliable} for those who need robot services, as shown in Fig. \ref{motivationfigure}.

Language-based solutions have an increasing potential in embodied AI applications, including tasks such as imitation learning, planning, control, and HRI \citep{jeon2024deep, zhao2023chat, c7}. However, relying on large language models (LLMs) for object perception often requires a supervised model and an understanding of semantic meaning \citep{esfahani2019deepdsair,liao2023se,yin2023segregator,ji2024sgba,yin2024outram}, which poses challenges when encountering rare or unknown objects. LLMs also have other problems, like hallucination for untrained tasks and tedious syntax structures. Despite these challenges, they serve as inspiration for new research efforts.

To address the challenge of unknown objects and improve HRI intuitiveness, we propose NVP-HRI, a zero-shot \textbf{n}atural multi-modal HRI approach integrating \textbf{v}oice and \textbf{p}osture commands via LLM \citep{lai2025nmm}. NVP-HRI makes use of the pre-trained Segment Anything Model (SAM) \citep{kirillov2023segment} to select areas of interest and enable zero-shot inference of point clouds for unknown objects. Our method also incorporates an LLM to compile actions and goals, ensuring collision-free real-time trajectory solutions. The generated action sequences are subject to further cross-check for content accuracy and safety, enabling efficient construction of complex control sequences.

Our main contributions are summarized below:
\begin{itemize}
\item 
We propose NVP-HRI, a multi-modal HRI framework for robot manipulators, efficiently segmenting zero-shot unknown objects and fusing them with parallel multi-modal inputs to construct complex temporal control sequences and references.
\item  
Our system utilizes an LLM to generate robot control sequences, incorporating language, posture, and zero-shot inference input. This effectively prevents collisions without human intervention and mitigates semantic errors. We achieve this by carefully structuring input representations and output tokens to address hallucination issues of LLMs, ensuring both safety and performance.
\item 
We benchmark our system against state-of-the-art HRI methods, showing strong performance with minimal syntax memorization and rapid input speed. Our system will be open source on \url{https://github.com/laiyuzhi/NVP-HRI.git}.
\end{itemize}

\section{RELATED WORK}

 Object detection \citep{wang2015vision} plays a crucial role in HRI systems, enabling robots to interact accurately with the world \citep{li2024jacquard}. As the standard of living increases, the requirements for object detection also need to increase, particularly in zero-shot rare object detection \citep{esfahani2021learning}. Existing systems predominantly rely on pre-trained supervised networks, such as Mask R-CNN \citep{hameed2022score}, YOLO \citep{li2023yolo}, and SSD \citep{wang2022slms}. The training of these networks relies on a large number of annotated datasets such as ImageNet \citep{deng2009imagenet}, Microsoft COCO \citep{lin2014microsoft}, Jacquard \citep{depierre2018jacquard, li2024jacquard} and MCD \citep{nguyen2024mcd}. However, these supervised models have problems in dealing with rare \citep{cao2023mopa}, unknown or new objects \citep{7064275}. Some works \citep{9340893, 10.1007/978-3-319-10401-0_19} explored new modalities to deal with unknown objects \citep{zhou2023metafi}, such as using a head mounted eye tracker \citep{9340893} or a depth camera \citep{10.1007/978-3-319-10401-0_19}. Some researchers \citep{9340893} have implemented a distillation-based method using gaze with a head-mounted eye tracker to detect unknown objects. However, this method relies on the accuracy of the hardware as well as precise gazing at the target object for an extended period of time, which is inefficient. In addition, using and carrying an eye tracker can be a challenge for the elderly. Depth cameras \citep{ji2022robust,Xu2024M}, as the most commonly used hardware devices in HRI, also have capabilities in the detection of unknown objects. The fixation-based unknown object segmentation approach \citep{10.1007/978-3-319-10401-0_19} was proposed by combining RGB images and depth maps. Using the Graph Cut algorithm \citep{boykov2004experimental} as an intermediate step and refining it with the Grab Cut algorithm \citep{rother2004grabcut}, this approach demonstrates notable improvements in processing speed. However, despite these optimizations, the processing time requires more than two seconds per image.  Our proposed NVP-HRI uses pretrained SAM \citep{kirillov2023segment} to obtain point cloud and accurate representation of zero shot objects. Compared to HRI methods using supervised pre-trained models \citep{hanggesture, stepputtis2020language,ConstantinEYBW22}, our approach is more suitable for real-world challenges on object detection brought by the increase in living standards. Furthermore, our method does not require users to wear additional equipment \citep{deng2024compact,deng2024incremental}, enhancing usability and efficiency.

In recent years, developing efficient, intuitive and natural HRI methods has become a key research priority for future service robots. 
Some systems rely on singular modality input, such as gesture \citep{hanggesture}, tactile \citep{wang2024touch}, surface electromyography (EMG) \citep{zhang2024robust}  and audio input \citep{c14,HuangWZL0023}. However, gesture-based methods require the user to accurately memorize and execute gestures corresponding to the predefined commands, which is a challenge. The tactile input generally uses only the pressure sensor arrays. This allows the user to express only a relatively simple intention with low accuracy. EMG signals can only predict simple human intentions and cannot perform complex human intentions, such as clearing the desktop. The audio input \citep{yang2023av,yang2024av} is prone to ambiguity when describing complex scenarios \citep{yuan2024MMAUD} because sentences can be interpreted in multiple ways. To improve the intuitiveness of HRI, we propose the NVP-HRI system, which combines the strengths of posture and natural language. This multimodal approach allows users to select objects with deictic posture, avoiding the need for detailed or ambiguous descriptions, while commands are given in natural language, eliminating the necessity to memorize complex gestures.

Multimodal HRI approaches \citep{wang2017heterogeneous, deng2022gaitfi,zhou2023metafi,yang2024mm} show great potential to solve these issues. 
Fusion of gesture and facial expression \citep{li2023mmfn} is proposed to categorize human emotions for HRI. However, simple fusion is not capable of understanding complex human intentions. 
However, simple fusion methods are not sufficient for understanding complex human intentions. In recent years, wearable mixed reality (MR) devices, such as Apple Vision Pro, have shown promise as multimodal HRI input devices \citep{li2024hcto,li2024helmetposer,park2024self,nguyen2023vr,yang2024fast}. Despite their potential, these devices come with steep learning curves, heavy weights, and high costs, making them unlikely to be widely adopted in elderly care and medical applications. In contrast, our proposed NVP-HRI system is based solely on a microphone \citep{lei2025audio} and a depth camera \cite{Xu2024M}, eliminating the need for users to wear additional devices. This makes it more accessible and suitable for everyday use, particularly in settings of elderly care and patient assistance.

 \begin{figure*}[t]
      \centering
      \includegraphics[width=0.9\textwidth]{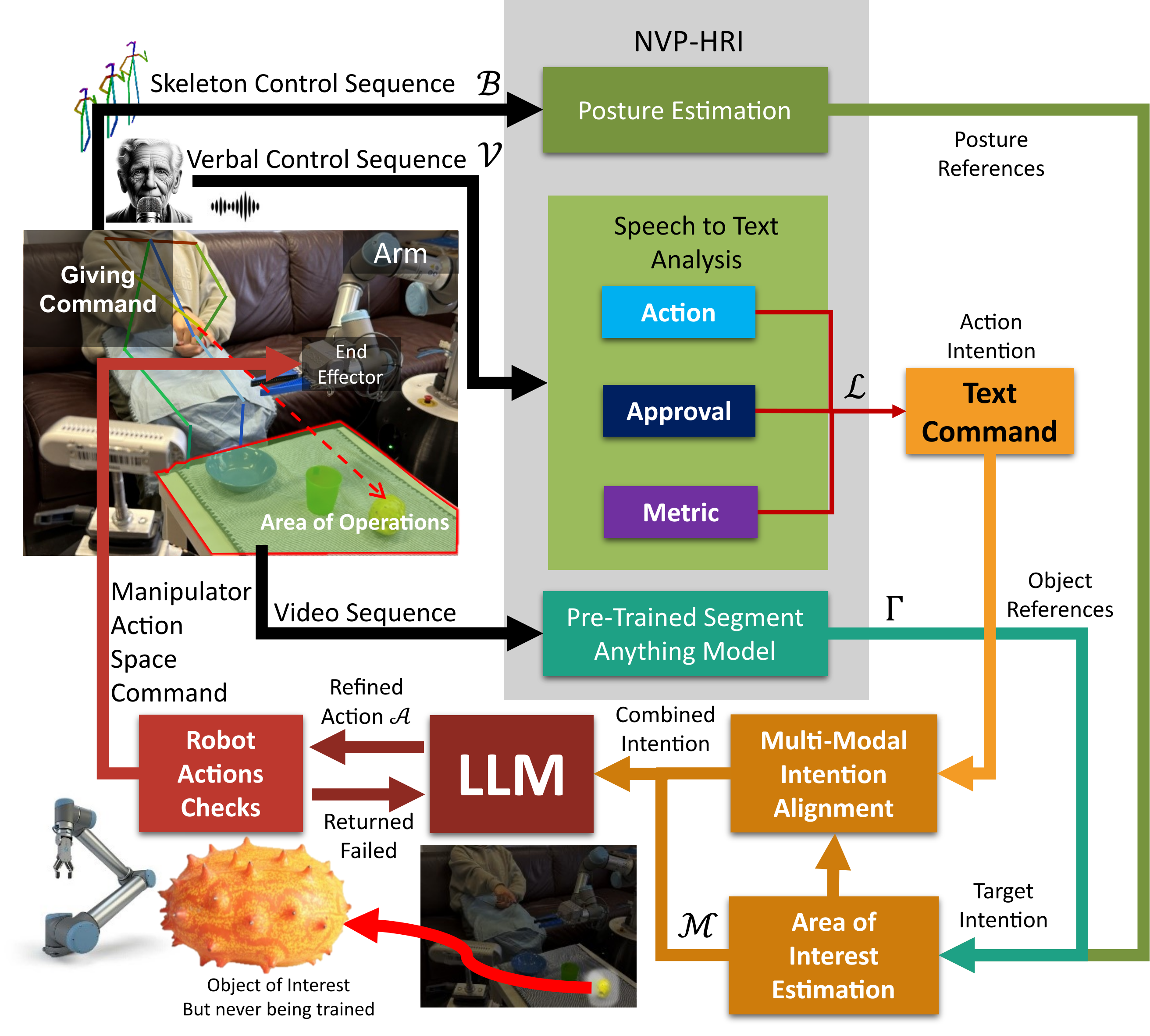}
      \caption{System Overview.
$\mathcal{V}$ represents verbal command, $\mathcal{B}$ represents posture references, $\mathcal{L}$ is mapping vocal features to text, $ \Gamma$ represent the 3D cluster of the scene object, $ \mathcal{M}$ is the mapping to get the target intention, $ \mathcal{A}$ is the mapping to get the action sequences.}
      \label{systemdiagram}
   \end{figure*}   
Although robotics in the medical field often focuses on surgical assist robots \citep{bian2023learning, mahmood2022dsrd} and rehabilitation robots \citep{shah2024efficient}, these applications face stringent safety requirements and the need to handle complex and variable situations \citep{li2024gera}, delaying widespread adoption. Service robots for elderly care and patients, on the other hand, show more immediate potential.
Current systems for the elderly or patients often require complex gestures or verbal commands due to ineffective information integration. Leveraging LLMs with robust reasoning offers promise for enhancing multimodal data processing. It can be applied to variable aspects of HRI \citep{zhang2023large}, including social robots with specialized knowledge \citep{caruccio2024can, khennouche2024revolutionizing} and robot task planning \citep{10161317}. However, there is still a large gap between innovations and practical applications \cite{cao2020online}. LLMs can generate human-like responses, but they can also generate inappropriate or even harmful content or actions due to the hallucination problem. Recently, we have seen many applications \citep{HuangWZL0023} that utilize pre-trained LLM with VLM to understand and interact with the environment \citep{jin2024robust,Li2024PSS}. However, it is generally based on language-only solutions. LLM-based posture and language fusion HRI models \citep{ConstantinEYBW22} still need improvements to understand the context in 3D spaces and manage interactions and behaviors. 
In our proposed system, we address these limitations by selecting the Segment Anything Model (SAM) \citep{kirillov2023segment}  over other unsupervised segmentation models due to its superior ability to extract detailed structural object representations and ease of implementation. This, combined with depth camera data, allows the LLM to better understand 3D scenes \citep{esfahani2019towards,ji2024lio,chen2024salient,ma2024mm} and plan collision-free trajectories based on human intentions.

When it comes to trajectory \citep{xu2024cost,Bai2024MRA,huswept2024} and action generation, traditional approaches have focused on methods such as preset greedy search \citep{Garrett2018PDDLStreamIS, feng2024adaptive}, behavior trees \citep{c8}, deep learning-based solution \citep{wu2018learn,wu2019tdpp,wu2019bnd,wu2019depth,wu2021learn} or Bayesian inference \citep{c18,cao2022direct}.
Recently, the integration of LLMs \citep{c7} into the generation of robot actions \citep{qi2024air} was examined, establishing a pipeline to translate human intentions into robot action sequences using an LLM. However, LLMs face limitations in directly accessing sensor data, while Vision Language Models \citep{ConstantinEYBW22} often require significant computational resources to generate sets of 2D embeddings \citep{liu2025handle}. These embeddings frequently contain errors and uncertainties, especially when encountering previously unseen objects. Furthermore, LLMs are prone to generating incorrect responses due to hallucination problems. Consequently, such LLM+VLM systems are prone to errors and mistakes, often requiring multiple trials to achieve success. To generate reliable action sequences, the symbolic action sequences generated by LLM are checked by visualizing the trajectory. To minimize hallucination, we formulate prompts and impose constraints on the output response tokens generated by LLM.


\section{PROBLEM DEFINITION}
\label{sec:PROBLEM FORMULATION}
An overview of our proposed NVP-HRI is shown in Fig. \ref{systemdiagram}. 
Our purpose is to create parallel multi-modal commands for rare or unknown objects, utilizing SAM to derive structural objects representation (See Fig  \ref{segment}). These commands and representations are then processed to generate real-time collision-free solutions via a token-constrained LLM.
\subsection{System Overview}
\textbf{Structural objects representation:} 

We represent the objects in the scene by classifying all objects as 3D point cloud clusters denoted $\mathcal{P} \in \mathbb{R}^3$. We perform a structural representation of each cluster denoted as $\Gamma(\cdot) \in \mathbb{R}^7$ including cluster index $i \in \mathbb{R} $, cluster height $h \in \mathbb{R}$, cluster width $w \in \mathbb{R}$, cluster thicknesses $d \in \mathbb{R}$, and 3D cluster centroid position $\gamma \in \mathbb{R}^3$. $q(t)\in \mathbb{R}^7$  denotes the state of the robot end effector at time t, including the 6D pose and the opening angle of the gripper. Consequently, the environment tuple $\mathcal{E}$ that includes the robotic arm manipulator and structural representation of the objects can be formed by $\mathcal{E}=(\Gamma(i, h, w, \gamma),\ q(t))$.



\textbf{Intention definition:} In order for LLM to control the manipulator $q(t)$ using structural object representation $\Gamma$, the complete human intention $I$ is inferred from a set of sparse keywords $\mathcal{S}$, which is decomposed into target references $I_t$, action reference $I_a$ and metric references $\lambda$. Our proposed NVP-HRI system utilizes RGBD and audio sensors to generate verbal input commands $\mathcal{V}$ and human posture $\mathcal{B}$.
Target references $I_t$ identify the specific object with which a human intends to interact based on $\mathcal{B}$. The action reference $I_a$ indicates the anticipated interaction with the target object based on $\mathcal{V}$. Metric references $\lambda$ serve to enhance input verbal information based on $\mathcal{V}$.

\subsection{Parallel Multi-modal Commands}
Our multi-modal HRI includes verbal input commands $\mathcal{V}$ and human posture $\mathcal{B}$ . The detailed meaning of the list of multimodal command sequences can be found below. 
\subsubsection{Verbal Command Sequence} 
\label{sec:verbal command}
Verbal language input is used to compile a set of commands and metrics due to its intuitive and naturalness. We divide $\mathcal{V}$ into three types:
\begin{itemize}
    \item Action command: designated for prospective actions.
    \item  Approval command: used to confirm the location of the selected object. 
    \item Metric references: designed as an optional part of human intention $I$. This command could include variables such as the angle of tilt for pouring or movement velocity.
\end{itemize}
The mapping $\mathcal{L}$ converts all verbal inputs into text and executes a query task to obtain the action intention $I_a$ and the metric $\lambda$. $\mathcal{L}$ is defined as:
\begin{equation}
    (I_a,\ \lambda) = \mathcal{L}(\mathcal{V})
\end{equation}




\subsubsection{Deictic Posture}  Deictic posture is a specific type of human posture used to indicate or point to an object, location, or direction. It involves gestures such as pointing with a finger, hand, or arm to direct attention and provide context in communication. NVP-HRI utilizes the RGBD input to generate the human postures represented by $\mathcal{B}$. The mapping to obtain the target intention based on the representation of the structural object $\Gamma$ from the human postures is denoted as:

\begin{equation}
    I_t = \mathcal{M}(\mathcal{B},\ \Gamma)
\end{equation}
The extended line $\ell$ of the right forearm of an operator extracted from $\mathcal{B}$ represents the deictic posture and the object closest to the vector $\ell$ as the target intention $I_t$. 
The distance of an object in the structural representation $\Gamma^k \in \Gamma$ with the centroid position of the 3D groups $\gamma^k$ to the vector $\ell$ is described as $d^k(l, \gamma)$:
\begin{equation}
d^k = \frac{\left | (\ell_2-\ell_1)\times (\ell_1-\gamma^k) \right | }{\left | \ell_2-\ell_1 \right | }   \label{eq:dis}
\end{equation}
Where $\ell_1$ and $\ell_2$ are two random points from the vector $\ell$. 

By integrating the multi-modal HRI command sequence $\mathcal{S}$ and the structural object representation $\Gamma$, we can have good estimate of the human intention $I$, which encompasses the object intention $I_a$, action intention $I_o$, and metric parameter $\lambda$:
\begin{align}
    I  &\triangleq \mathcal{H}(\mathcal{S}, \ \Gamma) \\\label{eq:multimodal command} 
    &= \mathcal{H}(I_a, \ I_t, \ \lambda) \notag
\end{align}


In our proposed system, we employ GPT4 \citep{c23} to decode a multimodal command sequence $\mathcal{S}$ into human intention $I$, denoted by mapping $\mathcal{H}$. The robot action sequences $a$ are derived from the intention of humans $I$ through the mapping $\mathcal{A}: a = \mathcal{A}(I)$. This mapping is also performed utilizing GPT4. Finally, the action sequences cross-checked by the swept volume model and our proposed system are used to control the state of the end effector $q(t)$ to fulfill the human intention $I$.


\subsection{Construction of Complex Commands}

Various action commands require various specifications for the intention of the target. Simple tasks specify only the intention of the action and have no target intention (e.g., {move to the initial position}). For a human intention such as \textit{pick up this exotic fruit}, both the action and the target intentions are necessary. In home service scenarios, human action intentions often occur in sequences, such as \textit{pick up the exotic fruit and give it to me}. In order to achieve this goal, any two robotic actions in our multimodal HRI can be constructed together:

\begin{equation}
     \mathcal{S}  = (\mathcal{S}_1, \mathcal{S}_2) = (I_{a1},\ I_{t1}, \ I_{a2},\ I_{t2},\ \lambda) \label{eq:multimodal command2} 
\end{equation}
Here, $\mathcal{S}_1$ and $\mathcal{S}_2$ denote the individual syntax commands. $I_{a1}$ and $I_{a2}$ denoted the intention of action of each syntax command. $I_{t1}$ and $I_{t2}$ denoted the target intention of each syntax command.
NVP-HRI comprises a series of robotic actions characterized by increasing complexity:
\begin{itemize}
    \item Without a clear goal: reset position, drop, move, etc.
    \item With targeted intention: pick, place, pour, etc. 
\end{itemize}

 \begin{figure*}[Ht]
      \centering
      \includegraphics[width=0.88\textwidth]{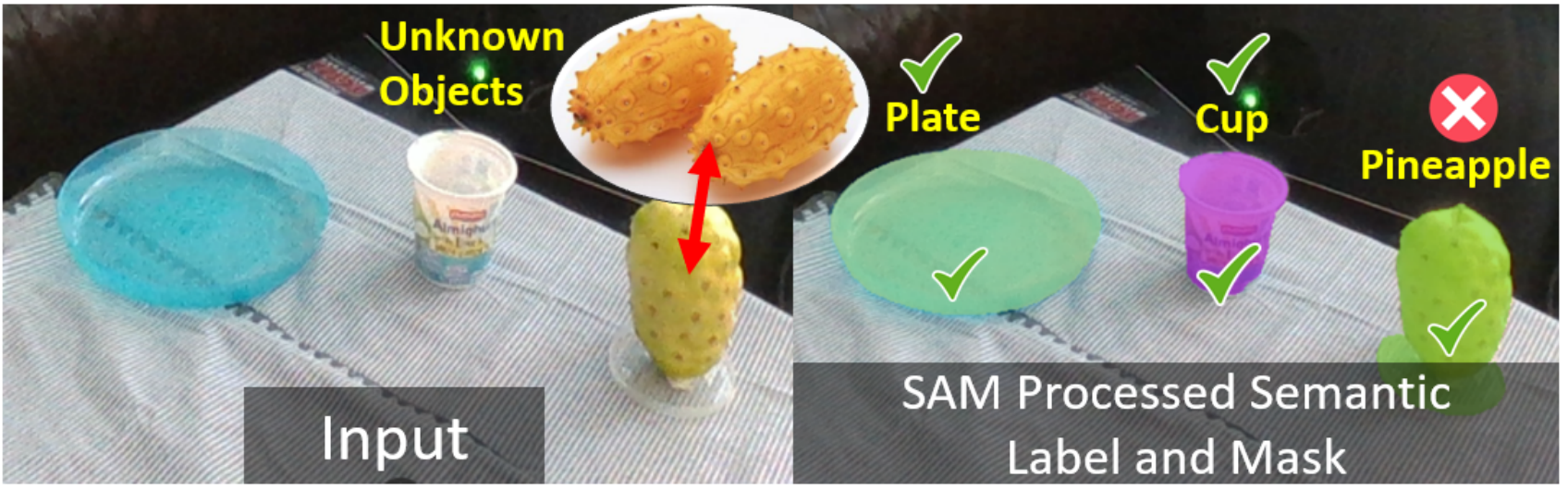}
      \caption{ Input and Output of SAM: Out of three objects, only two have correct semantic meanings. However, all of them are segmented correctly. }
      \label{segment}
   \end{figure*}    
\section{Methodology}
\label{section:MATERIALS AND METHODS}

Our application scenario involves indoor service for the elderly, where the robot must perform tasks ordered by humans efficiently and naturally. Some objects in this scenario are rare and may not be recognized by pre-trained models. To address this, NVP-HRI includes various modules, such as verbal command understanding, zero-shot semantic segmentation by pretrained SAM, deictic posture estimation, action sequence generation, and execution.
  
\subsection{Verbal Commands Understanding}

In the verbal commands understanding module, the intention and action metric are estimated from the raw audio input $\mathcal{V}$. This mapping $\mathcal{L}$ consists of two steps. The first step is the conversion of speech to text. The second step involves a query task to recognize the intention and the action metric.

After comparing various approaches \citep{c24,c25,c26}, we decided to utilize the VOSK API \citep{c24} based on the following criteria: (1) Recognition accuracy, (2) Recognition efficiency, and (3) Capability to identify individual speakers. 

\subsection{Zero-Shot Semantic Segmentation}

The semantic segmentation module extracts the representation of structural objects in the robot base frame by utilizing pre-trained SAM \footnote{\textcolor{black}{Checkpoint: \texttt{vit\_l}, Input Size: $1024 \times 1024 \times 3$}} \citep{kirillov2023segment}.
As shown in Fig. \ref{segment}, the extracted semantic mask may contain incorrect semantic meanings from the zero-shot prediction, which we will ignore and will never be used in the system. Instead, the texture and appearance clustering and grouping should remain relatively stable to provide rough estimates of object contours. In a random scene, we may have as many 2D clusters as possible. We then use the default calibration values $\mathcal{K}$, stored in the RGBD camera, to generate the reprojection matrix to project the RGBD 2D pixel $p$ into 3D point clouds ${ }^{C} \mathcal{P}$ through $  { }^{C} \mathcal{P} = \mathcal{K}p$. Thus, sets of 3D cluster candidates can be obtained in the camera frame. Here, we limit the size of interactable objects to be within the width of the manipulator end effector for interactions. Larger objects are stored for further scene understanding but are not used for reference object selection. Subsequently, the transform matrix ${ }^{B} \mathcal{T} _{C}$ \citep{34770} is performed to obtain the 3D point cloud of the object in the robot base frame through ${ }^{B} \mathcal{P} ={ }^{B} \mathcal{T} _{C}{ }^{C} \mathcal{P}$. 
Where, $ { }^{C} \mathcal{P}$ denoted a point in the point cloud in the camera frame and $ { }^{B} \mathcal{P}(.)$ denote a point in the point cloud in robot base frame represented as $ { }^{B} \mathcal{O}(x,y,z)$. The object position $\gamma$ in the scene is outlined as the centroid of the object point cloud in the robot base frame:

\begin{equation}
 \gamma^k = (\frac{1}{N} \sum_{i=1}^{N}x_i^k ,\ \frac{1}{N} \sum_{i=1}^{N}y_i^k,\ \frac{1}{N} \sum_{i=1}^{N}z_i^k) \label{eq:xi}
\end{equation}
Where $(x_i,y_i,z_i)$ denotes the coordinates of a random point with index $i$  in the point cloud of object $k$ in the robot base frame. The \textcolor{black}{3D clusters centroid position} $\gamma^k$ is then used to calculate the distance, as described in Eq. \ref{eq:dis}. The object width $w$ , thickness $d$, and the object height $h$ are defined as:
\textcolor{black}{\begin{align} 
    w  & = \max x^k -\min x^k \\\label{eq:cluster} \notag
    h  & = \max y^k -\min y^k \\ \notag
    d  & = \max z^k -\min z^k  \notag
\end{align}}
The width, denoted by $w$, will be used for the end effector opening, while $w$,  \textcolor{black}{$d$} and $h$ together are used for the path planning of the end effector \textcolor{black}{and the cross-check of the trajectory.}

\subsection{Deictic Posture Estimation}

This module is utilized to infer the intended object of interest. To ensure the system functionality, the upper half of the human body must be captured with an RGBD camera at a sufficient distance, resulting in a larger area of interaction compared to a Leap Motion-based solution. This can be considered as a hidden benefit of our approach. After that, we utilized OpenPose \citep{c12} to track 2D human skeletons within the RGB image. Subsequently, the aligned depth map is used to estimate the 3D human skeletons $\mathcal{B}$ in the camera frame. The direction line $\ell$ of the extended right forearm of the operator is then chosen as the direction of the deictic posture.

The intended object of interest can be computed in each frame in real-time. However, the relevant location becomes accessible only when the verbal approval command is identified. The precision and robustness of the deictic posture are evaluated in the following chapters.

\subsection{Action Sequence Generation and Execution} 

   
The translation of human intention $I$ into a sequence of actions $a$ for a robot is achieved by the mapping function $\mathcal{A}$. In the context of collision-free action sequence generation, an extensive understanding of the operational environment and the mission itself is required. 
\textcolor{black}{
To facilitate this process naturally,} we need a large language model for backend processing. To do that, we select \textcolor{black}{GPT-4-turbo \footnote{ \textcolor{black}{We use GPT-4-turbo with version 2024-04-09 and the training data retrieved in December 2023.}}} \citep{c23} for several reasons: its ability to encode high-level environmental representations, its ability to process complex natural language commands, its 3D spatial understanding, and its ease of integration due to user-friendly API interfaces and extensive pre-built samples. These features make \textcolor{black}{GPT-4-turbo} highly adaptable for a variety of scenarios. Acting as the central processor, GPT-4 handles multi-modal inputs and generates robot action sequences efficiently, ensuring smooth, collision-free operation aligned with human intentions. 

In order to address the potential issues of hallucination as well as tedious syntax structures in LLM, we formulate prompts and impose constraints on the output response tokens generated by the LLM. The prompt, as shown in Fig. \ref{actionsequence}, is structured into three separate components:

\begin{enumerate}
    \item  Action constraints: restricts the usages of actions. LLM is limited to utilizing these actions for constructing robot action sequences.
    \item  Trajectory constraints: limit specific requirements for the planning of the real-time trajectory, where it is necessary to define the units and the coordinate axes. 
    \item Example tasks: illustrate the execution of similar tasks and provide guidance on task planning strategies using LLM. By following this constraint, LLM will replicate the example task.
\end{enumerate}

\textcolor{black}{As shown in Fig \ref{actionsequence}, the GPT-generated plan first reconstructs the structural representation of the scene to determine the positions, widths, and heights of the manipulated objects and obstacles. Subsequently, the safe height is established based on trajectory constraints, and finally, the action sequences are generated according to action constraints and trajectory constraints, forming a trajectory awaiting cross-check mechanism. This trajectory needs to be able to avoid collisions. The format of the action sequences is determined by the examples in the prompt to ensure uniformity.}

\subsection{Cross-Check Mechanism}
To plan a collision-free trajectory, LLM also needs to incorporate the spatial relations of objects in the scene through the structural representation of the objects. To further enhance the safety and accuracy of NVP-HRI, the generated action sequences are subsequently cross-checked. \textcolor{black}{The decision to lift an object or move it horizontally is largely determined by GPT, which operates as a black-box system with embedded common sense reasoning. In the demo given in Fig. \ref{systemdiagram}, GPT likely chose to lift the object to avoid the obstacle because other options, such as moving in front of or behind the obstacle, might pose risks of potential collisions, such as with occluded objects or swept-area collisions. Decision making is context-dependent and fully depends on common sense. In other scenarios, the GPT might choose a sloped movement to maintain observability of the system. This demonstrates that GPT, as a pre-trained black-box model, possesses an inherent level of common sense and can adapt to different situations effectively.}

\begin{figure*}[Ht]
      \centering
      \includegraphics[width=0.895\textwidth]{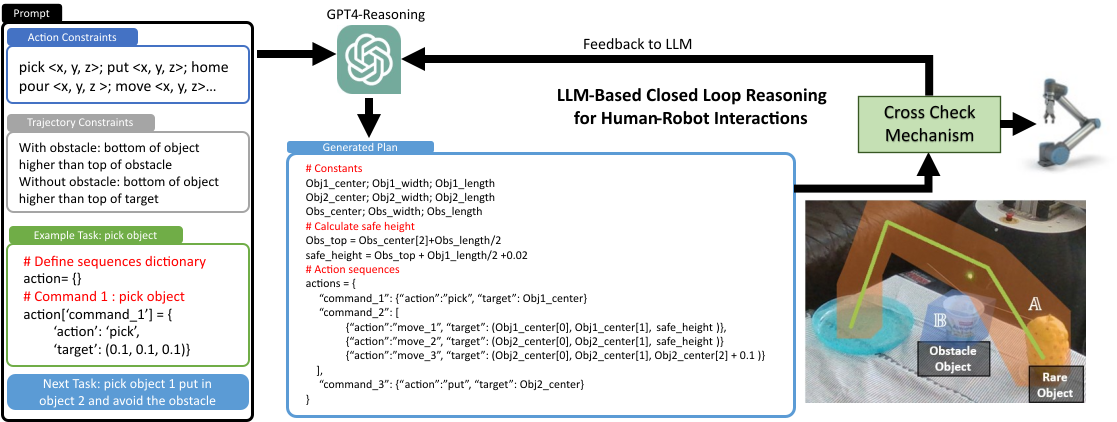}
      \caption{\textcolor{black}{The prompt is divided into action constraints, trajectory constraints, and example tasks, followed by a cross-check. Cross-check results are fed back to the LLM, and if a collision is detected, a new trajectory is generated. Only sequences that pass the cross-check are executed by the robot, resembling a closed-loop control system as in classical control theory.}}
      \label{actionsequence}
\end{figure*}

Hence, in our proposed system, the object to be manipulated is restricted by the trajectory constraints embedded in the prompts along the generated trajectory to cross obstacles from above. In the cross-check process, the manipulated object and obstacle are simplified to two rectangular shapes based on their structural representations $\Gamma^M$ and $\Gamma^O$. $\mathbb{A} = \cup_{a\in[a_s,a_e]}\Gamma^M\psi (a)$ represents the swept volume of the manipulated object along the trajectory $\psi$ from the start action sequence $a_s$ to the end action sequence $a_e$, as indicated by its structural representation $\Gamma^M$. $\mathbb{B}=\Gamma^O$ denotes the volume of the obstacle, as determined by the structural representation $\Gamma^O$. If $\mathbb{A}\cap \mathbb{B}\ne \emptyset $, then there is a collision in the trajectory. \textcolor{black}{ Under this circumstances, GPT-4 provides feedback on the generated trajectory, including detailed explanations of failure causes, such as the exact location of a collision. This explanatory feedback helps GPT-4 adjust effectively in subsequent reasoning.} If $\mathbb{A}\cap \mathbb{B} = \emptyset $, then the trajectory is collision-free. As shown in Fig. \ref{actionsequence}, the visualization of the collision-free trajectory can be carried out efficiently by the proposed solution. After passing the cross-check, the robot executes the action sequences, resembling a closed-loop feedback system as in traditional control theory. \textcolor{black}{The cross-check mechanism significantly improves the accuracy of collision-free trajectory generation and ensures safer task execution, increasing the success rate of collision-free trajectories from 73\% to 94\% compared to direct trajectory generation. }


\subsection{Human-Robot Interaction}
\textcolor{black}{Human-robot interaction in our proposed NVP-HRI is accomplished by multi-modal command sequences, in which the user communicates intentions through verbal command sequences and deictic posture. Multi-modal commands are processed by $\mathcal{H}$, and visual feedback is provided through the graphical user interface to improve interaction accuracy and robustness.}

1) The user gives an action command by voice, which is estimated by $\mathcal{L}$ as the user action intention. 2) While commanding the voice, the operator also has to make a deictic posture giving the target object through mapping $\mathcal{M}$. With the approval command, the user determines the selection of the target object. 3) The user can optionally give a metric command, which is estimated by $\mathcal{L}$ as the metric parameter for the human intention.
In NVP-HRI, the users express intentions using verbal commands and deictic postures via process $\mathcal{H}$, while receiving visual feedback (see Fig. \ref{fig1:visualfeedback}) from the graphical user interface.

Transitioning from multi-modal command sequence to human intention involves:
\begin{enumerate}
\item User determines verbal action command. If needed, a user selects a target.
\item System calculates target object using deictic posture $\ell$ and 3D object position $\gamma$.
\item Optional metrics references $\lambda$ enhance robotic manipulation.
\item User confirms intention using \textit{finish} character.
\end{enumerate}

These are all verbal instructions given to the participants in the experiment.

\label{environment}
\section{EXPERIMENT}
\textcolor{black}{The experiment aims to validate the superiority of our proposed NVP-HRI. To achieve this goal, we divide our experiment into offsite survey approaches and onsite physical trials. We conduct field offsite surveys to engage as many people as possible to collect the expected user experience to reflect intuitiveness and efficiency. With a set of survey candidates, some agreed to our invitation and performed onsite tests to verify accuracy and precision. Thus, we can perform the quantitative analysis of the proposed HRI framework. Furthermore, we verified the robustness of our proposed method for interacting with unknown objects and conducted perception tests in a real-world environment in an elderly care center to showcase the qualitative value of the proposed HRI framework.}

\subsection{Environment Setup}
The experiment environment setup resembles a typical living room scenario. An Intel D435i RGBD camera is strategically placed to oversee the area, providing clear visibility of humans on the sofa, as well as the UR robot manipulator and table. \textcolor{black}{The RGB output of the camera is $1024\times768\times3$, and the depth map is aligned with the RGB image, resulting in an output of $1024\times768\times1$. To adapt to the input requirements of different models (e.g., YOLOv5 and SAM), the camera's output is resized to the corresponding dimensions.} At present, we are utilizing the microphone on the notebook for processing, as no dedicated audio device is available.

Our system has been tested with 20 participants of different age groups in this environment, as depicted in Fig. \ref{fig1:visualfeedback}. 
We instructed participants to act as if they were unable to move normally and to depend on the robot arm to transport items within their reach. The objective was for the robot arm to retrieve the necessary items for the participants. \textcolor{black}{When testing our proposed system, }some of these items, such as charcoal brioche and exotic tropical fruits, were not recognized by the pre-trained object detection model or humans. Overall, this creates a realistic environment for testing the HRI system.

\subsection{Description of Experimental Scenarios}
In the experimental scenarios, a series of manipulation experiments were devised to explore the increasing complexity of human intention, denoted as \\
$I = \mathcal{H}(I_{a1},\ I_{t1}, \ I_{a2},\ I_{t2},\ \lambda)$:
\begin{enumerate}  
\item[$\bullet$]$(home,\ -,\ -,\ -, \ -)$, $(throw,\ -,\ -,\ -, \ -)$
\item[$\bullet$]$(pick,\ unknown \ object,\ -,\ -, \ -)$
\item[$\bullet$] $(pick,\ unknown \ object,\ put,\ bowl,\ -)$
\item[$\bullet$] $(pick,\ cup,\ pour,\ unknown \ object,\ ang = 90 ^{\circ})$
\item[$\bullet$] \textbf{Multistep tasks:} add water and pass, pour muesli and add milk.
\end{enumerate}

\subsection{\textcolor{black}{Benchmarking Criteria}}
\textcolor{black}{
The primary objective is to measure the duration of user interaction with general objects and the success rate for each given scenario. The secondary objective is to evaluate the user-friendliness of the NVP-HRI. The performance of these scenarios is demonstrated in the provided videos. 
}\\
\noindent \textcolor{black}{ \textit{Remark:} To allow the existing method to work and to conduct fair benchmarking in efficiency and time analysis, we use a common recognizable item, such as cups and bowls. Otherwise, the existing method often cannot work with unknown items. The tests with rare objects are conducted separately, as other methods fail to recognize the objects.
To minimize order effects during baseline comparisons, we tested the accuracy and duration of the interaction of all participants between the baseline methods and our proposed method, employing different interaction orders. We conducted two-way ANOVA (Analysis of variance) to determine the effect of the order of interaction methods and the age group of the participants on the comparison of effectiveness and precision. }

\subsection{\textcolor{black}{Baseline Selections}} 

\begin{figure}[t]
\centering
\includegraphics[width=0.46\textwidth]{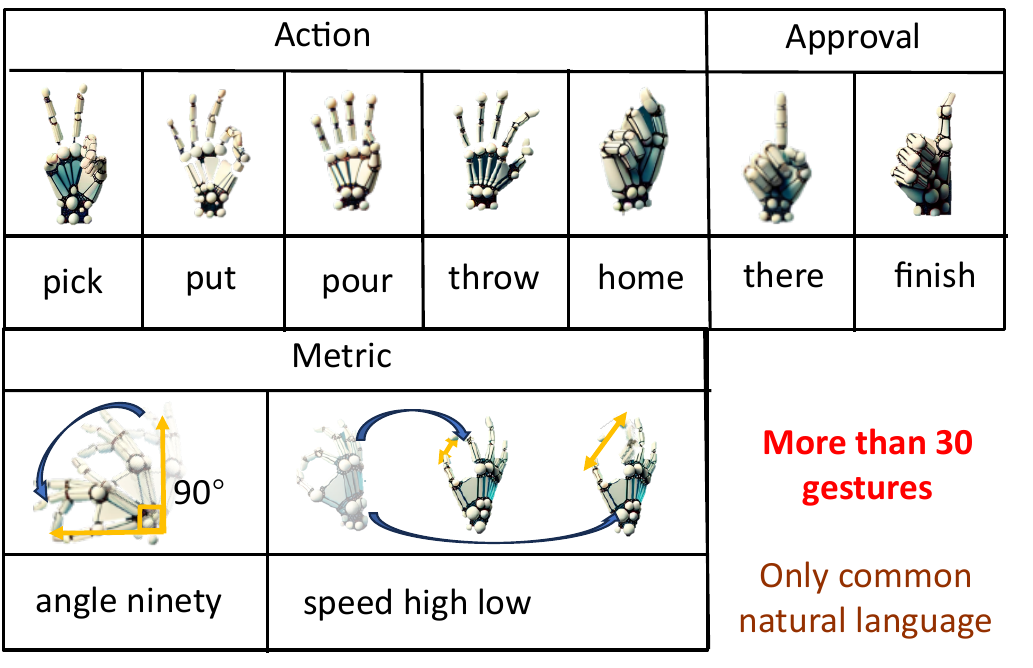}
\caption{Typical gestures utilized within gesture-based HRI system \citep{hanggesture} with their respective verbal commands.} 
\label{mapping}
\end{figure}

In our study, we benchmark our approach against other state-of-the-art gesture-based HRI methods \citep{hanggesture}, NLP \textcolor{black}{(Natural Language Processing)}-based methods \citep{stepputtis2020language}, and VLM \textcolor{black}{(Vision Language Model)}-based methods \citep{ConstantinEYBW22}. These baselines are chosen because all these methods have open source implementations and are somewhat similar to our approach. Most existing multimodal approaches \citep{c13, c16} do not have open-source source code for comparison, which prevents us from making direct comparisons. \textcolor{black}{The gesture-based \citep{hanggesture} utilizes a fixed position leap motion \citep{leapmotion} sensor to capture the structure of the hand bone. The NLP and VLM methods \citep{stepputtis2020language, ConstantinEYBW22} use natural language sentences to direct the actions of a robot. \textcolor{black}{NLP-based method \citep{stepputtis2020language} integrate word embedding, attention mechanisms, and probabilistic reasoning to recognize objects described in natural language, while VLM-based method \citep{ConstantinEYBW22} utilize the CLIP visual-language model \citep{radford2021learning} to resolve ambiguities of object selection and help users formulate clearer expressions.} However, based on feedback from participants in experiments and surveys, there are limitations to each of the compatible methods:
\begin{enumerate}
    \item The gesture-based method requires users to remember complex gestures compromising HRI intuitiveness, as shown in Fig. \ref{mapping}. Those are the gesture commands that we ask
the participant to memorize
    \item The Leap Motion's fixed position and limited field of view require the user to maintain their hand above the sensor, posing challenges for all participants.
    \item The natural language sentences struggle with distinguishing between similar objects in different locations and with unknown objects.
    \item VLM-based methods require an interactive dialog in order to select a target object and are, therefore, slow and unable to discriminate between the same objects.
\end{enumerate}}

\textcolor{black}{Additional, the object detection modules of the baseline methods we chose for comparison are based on supervised models (e.g., FRCNN \footnote{\textcolor{black}{Network: Resnet101, Input Size: $1024\times 768\times 3$}} \citep{ren2016faster} and \textcolor{black}{YOLOV5} \footnote{\textcolor{black}{Checkpoint: YOLOv5l, Input Size: $640\times 640\times 3$}} \citep{li2023yolo}). Methods relying on the precise pronunciation of objects cannot handle unknown objects. Therefore, for baseline comparisons, we compare the duration and accuracy of the interaction using common objects.}

\section{RESULTS AND DISCUSSION}
\label{RESULTS}

\textcolor{black}{We evaluated the performance of our proposed NVP-HRI through a series of challenging experiments designed to assess criteria such as location robustness, accuracy, efficiency, and user experience.}



\begin{figure}[t]
      \centering
      \includegraphics[width=0.49\textwidth]{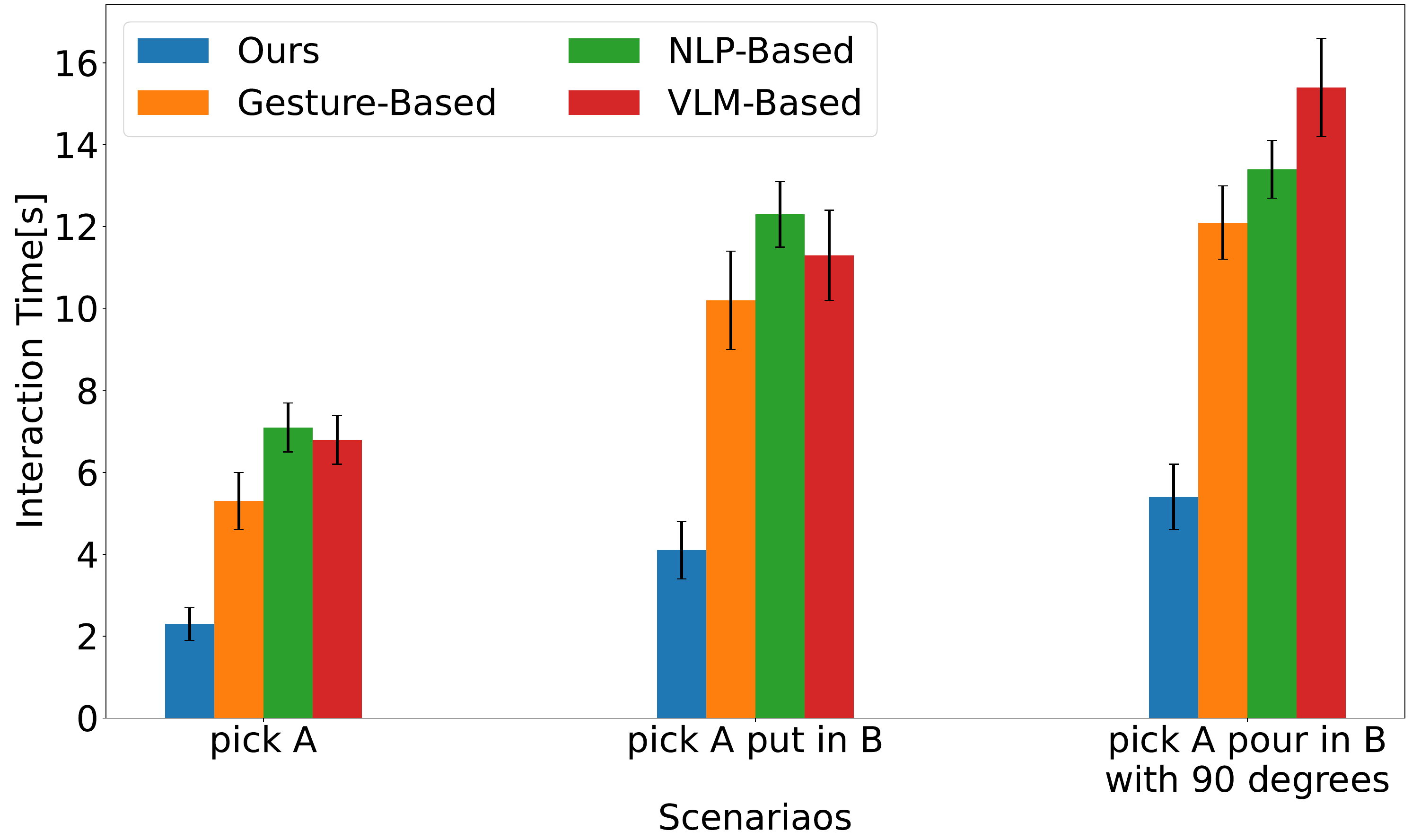}
      \caption{Duration of user interaction to enter the same commands as the gesture-based HRI and NLP-based HRI, our system required less time compared to hand gesture-based HRI and NLP-based HRI.}
      \label{episode}
\end{figure}
   
\subsection{Comparison of Effectiveness and Intuitiveness} 

\begin{table*}[thp]
\caption{\textcolor{black}{Interaction Duration in Different Scenarios for Different Age Groups.}}
\label{table_example4}
\begin{center}
\begin{tabular}{ccccccccccccccccccc}
\hline
\hline
\multicolumn{1}{c}{\multirow{3}{*}{Method}} & \multicolumn{9}{c}{Interaction Duration ($s$) $\downarrow$}  \\ \cmidrule(r){2-10} 
 & \multicolumn{3}{c}{$\mathbb{Y}$}  & \multicolumn{3}{c}{$\mathbb{M}$} & \multicolumn{3}{c}{$\mathbb{O}$}  \\ \cmidrule(r){2-4} \cmidrule(r){5-7} \cmidrule(r){8-10} 
 & S1 & S2 & S3 & S1 & S2 & S3 & S1 & S2 & S3  \\ \hline
  \shortstack{\\Gesture \\ \citep{hanggesture}} & \shortstack{\\4.93 $\pm$\\ 0.84}& \shortstack{\\9.61 $\pm$\\ 0.96}& \shortstack{\\11.67 $\pm$\\ 0.44}& \shortstack{\\5.13 $\pm$\\ 0.32}& \shortstack{\\9.89 $\pm$\\ 0.90}& \shortstack{\\11.75 $\pm$\\ 0.71}& \shortstack{\\5.92 $\pm$\\ 0.82}& \shortstack{\\11.31 $\pm$\\ 0.40}& \shortstack{\\13.03 $\pm$\\ 1.00}  \\ \hline
   \shortstack{\\NLP \\ \citep{stepputtis2020language}} & \shortstack{\\6.63 $\pm$\\ 0.41}& \shortstack{\\11.73 $\pm$\\ 0.71}& \shortstack{\\13.11 $\pm$\\ 0.76}& \shortstack{\\6.87 $\pm$\\ 0.60}& \shortstack{\\12.21 $\pm$\\ 0.30}& \shortstack{\\13.11 $\pm$\\ 0.86}& \shortstack{\\7.97 $\pm$\\ 0.54}& \shortstack{\\13.11 $\pm$\\ 0.91}& \shortstack{\\14.07 $\pm$\\ 0.40} \\ \hline
    \shortstack{\\VLM\\ \citep{ConstantinEYBW22}} & \shortstack{\\6.28 $\pm$\\ 0.53}& \shortstack{\\10.94 $\pm$\\ 0.89}& \shortstack{\\15.03 $\pm$\\ 0.81}& \shortstack{\\6.62 $\pm$\\ 0.64}& \shortstack{\\10.87 $\pm$\\ 0.52}& \shortstack{\\14.67 $\pm$\\ 0.52}& \shortstack{\\7.66 $\pm$\\ 0.33}& \shortstack{\\12.22 $\pm$\\1.07}& \shortstack{\\16.62 $\pm$\\ 0.78}  \\  \hline
    \shortstack{\\\textcolor{black}{NVP-HRI}\\ (Ours)}  & \shortstack{\\\textbf{2.55} $\pm$\\ 0.49} & \shortstack{\\\textbf{4.23} $\pm$\\ 0.99} & \shortstack{\\\textbf{5.12} $\pm$\\ 0.88} & \shortstack{\\\textbf{2.08} $\pm$\\ 0.80} & \shortstack{\\\textbf{3.80} $\pm$\\ 0.67} & \shortstack{\\\textbf{5.01} $\pm$\\ 0.80} & \shortstack{\\\textbf{2.17} $\pm$\\ 0.36} & \shortstack{\\\textbf{4.19} $\pm$\\ 0.64} & \shortstack{\\\textbf{6.14} $\pm$\\ 0.44}\\ 
\hline
\end{tabular}
\end{center}
\vspace{-10pt}
\footnotesize{  $\mathbb{Y}$ : younger age group of 18-35 y.o., $\mathbb{M}$: Middle-aged group of 36-50 y.o., $\mathbb{O}$: Older age group of 51-65 y.o.. }\\
\footnotesize{ Best results are in \textbf{bold}. The rest of the paper follows the same naming convention.}
\end{table*}

\begin{table*}[thp]
\vspace{-8pt}
\caption{\textcolor{black}{Two-Way ANOVA and Post-Hoc for Interaction Duration}}
\label{table_example1}
\begin{center}
\begin{tabular}{lccccccccccccccccccc}
\hline
\hline
\multicolumn{1}{c}{\multirow{3}{*}{Scenario}}& \multicolumn{1}{c}{\multirow{3}{*}{Method}} & \multicolumn{4}{c}{Two-Way ANOVA} & \multicolumn{6}{c}{Post-Hoc} \\ \cmidrule(r){3-6} \cmidrule(r){7-12}
& & \multicolumn{2}{c}{Age group}  & \multicolumn{2}{c}{Interaction order} & \multicolumn{2}{c}{$\mathbb{M}$ vs $\mathbb{O}$}  & \multicolumn{2}{c}{$\mathbb{M}$ vs $\mathbb{Y}$}  & \multicolumn{2}{c}{$\mathbb{O}$ vs $\mathbb{Y}$} \\ \cmidrule(r){3-4} \cmidrule(r){5-6} \cmidrule(r){7-8} \cmidrule(r){9-10} \cmidrule(r){11-12}
& & $F(2, 18)^{\dotplus}$ & $P^{\dagger}$  & $F(3, 18)$ & $P$ & $P_{adj}^{\circ}$ & reject$^{\circledast}$ & $P_{adj}$ & reject & $P_{adj}$  & reject \\ \hline
 \multicolumn{1}{c}{\multirow{4}{*}{S1}}&VLM &15.340&0.000$^{\ast}$&0.730&0.547&0.001&true&0.392&false&0.001&true \\ 
  &Gesture &4.059&0.035&0.513&0.678&0.089&false&0.838&false&0.028&ture \\ 
   &NLP &17.716&0.000$^{\ast}$&1.679&0.206&0.000$^{\ast}$&true&0.633&false&0.000*&true \\ 
    &\textcolor{black}{NVP-HRI} &0.713&0.556&1.411&0.269&0.939&false&0.258&false&0.417&false \\ \hline
     \multicolumn{1}{c}{\multirow{4}{*}{S2}}&VLM &0.599&0.623&6.144&0.009&0.011&true&0.982&false&0.017&true  \\ 
     &Gesture&9.702&0.001&0.467&0.708&0.005&true&0.755&false&0.000$^{\ast}$&true \\ 
      &NLP &1.507&0.246&9.144&0.002&0.039&true&0.340&false&0.001&true \\ 
      &\textcolor{black}{NVP-HRI} &1.411&0.269&0.712&0.557&0.939&false&0.257&false&0.416&false \\\hline
     \multicolumn{1}{c}{\multirow{4}{*}{S3}}   &VLM &0.234&0.971&15.543&0.000$^{\ast}$&0.000$^{\ast}$&true&0.582&false&0.000*&true  \\ 
       &Gesture &8.662&0.002&0.609&0.617&0.004&true&0.973&false&0.002&true \\ 
         &NLP &6.259&0.008&2.392&0.102&0.027&true&1.000&false&0.028&true \\ 
          
          &\textcolor{black}{NVP-HRI} &0.525&0.600&0.405&0.751&0.551&false&0.773&false&0.928&false\\ 
\hline
\end{tabular}
\end{center}
\vspace{-10pt}
\footnotesize{ $^{\dotplus}F(2, 18)$ measures the ratio of the variance explained by a factor to the residual variance. In $F(2, 18)$, the numerator degrees of freedom are $2$, the denominator degrees of freedom are $18$. 
$^{\dagger}P$ denotes the level of significant of the $F$-ratio}. $P<0.05$ indicates that the factor has a statistically significant effect.
$^{\circ}P_{adj}$ denotes the adjusted $P$-value, which accounts for multiple comparisons in post-hoc tests
\\
\footnotesize{$^{\circledast}reject$ denotes a decision rule based on the $P_{adj}$.$reject = false$ means no significant difference between the groups.
\\
$^{\ast}$ denotes $P<0.001$.}
The rest of the paper follows the same naming convention.

\end{table*}

\begin{table*}[thp]
\vspace{-8pt}
\caption{\textcolor{black}{Interaction Accuracy in Different Scenarios for Different Age Groups}}
\label{table_example6}
\begin{center}
\begin{tabular}{ccccccccccccccccccc}
\hline
\hline
\multicolumn{1}{c}{\multirow{3}{*}{Method}} & \multicolumn{9}{c}{Interaction Accuracy $\uparrow$ ($\% $)}  \\ \cmidrule(r){2-10} 
 & \multicolumn{3}{c}{$\mathbb{Y}$}  & \multicolumn{3}{c}{$\mathbb{M}$} & \multicolumn{3}{c}{$\mathbb{O}$}  \\ \cmidrule(r){2-4} \cmidrule(r){5-7} \cmidrule(r){8-10} 
 & S1 & S2 & S3 & S1 & S2 & S3 & S1 & S2 & S3  \\ \hline
  \shortstack{\\Gesture \\ \citep{hanggesture}} & \shortstack{\\100$\pm$\\ 0.0}& \shortstack{\\93.4 $\pm$\\ 0.8}& \shortstack{\\86.2 $\pm$\\ 2.5}& \shortstack{\\100$\pm$\\ 0.0}& \shortstack{\\93.0$\pm$\\ 1.5}& \shortstack{\\86.6 $\pm$\\ 9.4}& \shortstack{\\100 $\pm$\\ 0.0}& \shortstack{\\92.3 $\pm$\\ 1.7}& \shortstack{\\85.6 $\pm$\\ 1.3}  \\ \hline
   \shortstack{\\NLP \\ \citep{stepputtis2020language}} & \shortstack{\\97.2$\pm$\\ 1.2}& \shortstack{\\78.2 $\pm$\\ 1.2}& \shortstack{\\66.5 $\pm$\\ 2.7}& \shortstack{\\97.3 $\pm$\\ 1.4}& \shortstack{\\78.3 $\pm$\\ 1.8}& \shortstack{\\68.6 $\pm$\\ 2.2}& \shortstack{\\96.3$\pm$\\ 0.7}& \shortstack{\\77.3 $\pm$\\ 0.9}& \shortstack{\\67.6 $\pm$\\ 1.7} \\ \hline
    \shortstack{\\VLM\\ \citep{ConstantinEYBW22}} & \shortstack{\\96.3 $\pm$\\ 2.6}& \shortstack{\\41.7 $\pm$\\ 1.1}& \shortstack{\\34.2 $\pm$\\ 1.8}& \shortstack{\\95.6 $\pm$\\ 2.1}& \shortstack{\\43.3$\pm$\\ 2.2}& \shortstack{\\34.3 $\pm$\\ 1.3}& \shortstack{\\96.2 $\pm$\\ 2.1}& \shortstack{\\42.6 $\pm$\\ 0.9}& \shortstack{\\33.9 $\pm$\\ 2.3}  \\ \hline
     \shortstack{\\\textcolor{black}{NVP-HRI}\\ (Ours)}  & \shortstack{\\\textbf{100}$\pm$\\ 0.0}& \shortstack{\\\textbf{95.5} $\pm$\\ 0.8}& \shortstack{\\\textbf{90.9} $\pm$\\ 1.7}& \shortstack{\\\textbf{100}$\pm$\\ 0.0}& \shortstack{\\\textbf{95.1} $\pm$\\ 1.5}& \shortstack{\\\textbf{91.6} $\pm$\\ 1.3}& \shortstack{\\\textbf{100}$\pm$\\ 0.0}& \shortstack{\\\textbf{94.3} $\pm$\\ 1.3}& \shortstack{\\\textbf{91.3}$\pm$\\ 0.9} \\ 
\hline
\hline
\end{tabular}
\end{center}
\end{table*}

Our experimental scenario consisted of two cups with the same shape and different colors, two bowls with the same shape and different colors, and a plate. To evaluate the duration costs of user interaction, we asked participants to enter the same commands and intentions using different HRI approaches \textcolor{black}{in different orders}, including picking up one of the cups \textcolor{black}{(S1)}, picking up one of the cups and putting it on the plate \textcolor{black}{(S2)}, and picking up one of the cups and pouring it at a 90-degree angle into one of the bowls \textcolor{black}{(S3)}. \textcolor{black}{24} participants were recruited from the local university, \textcolor{black}{8} of whom were over 51 years old. All participants were fluent in English and received verbal instructions on the NVP-HRI without undergoing formal training or practice sessions. \textcolor{black}{Each participant completed scenarios S1 through S3 in a distinct order of interaction methods, repeating each method five times within each scenario.} The experiment results, depicted in Fig. \ref{episode}, show that our system consumes 59.2\%  less time than gesture-based HRI, 65.2\% less time than NLP-based HRI and 64.8 \% less time than VLM-based HRI. This experiment highlights the enhanced efficiency of NVP-HRI in interaction within a complex 3D environment.

\textcolor{black}{We categorized the participants into three different age groups: those aged 18-35 belonged to the younger group, those aged 36-50 belonged to the middle group, and those aged 51-65 belonged to the older group. Table \ref{table_example4} shows the mean duration of interaction for participants in different age groups to enter the same command in different scenarios with different HRI methods.}

\begin{figure*}[t]
\centering
\includegraphics[width=0.8\textwidth]{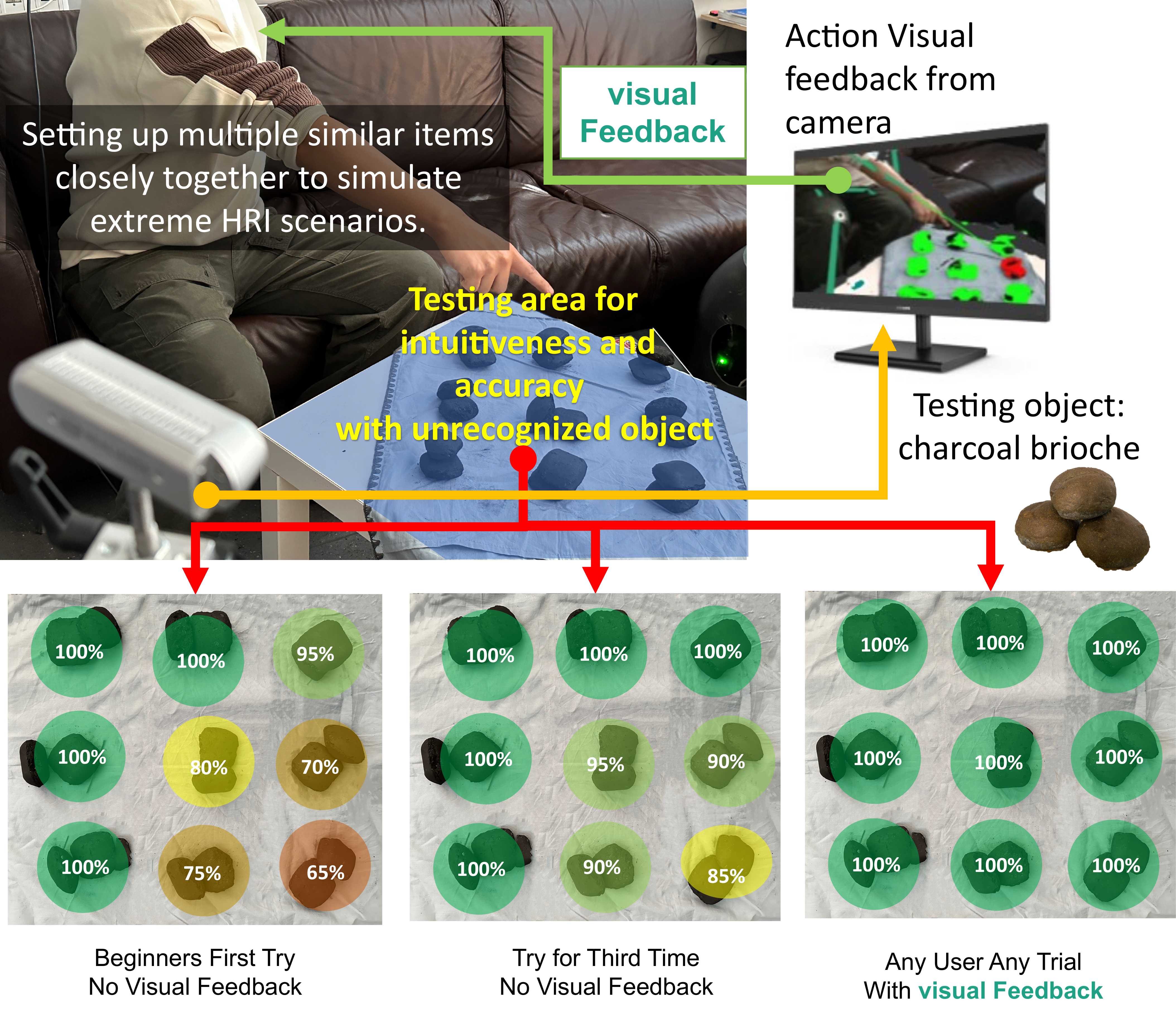}
\caption{Experiments to test the accuracy of deictic posture (above) and the deictic posture evaluation results (bottom).} 
\label{fig1:visualfeedback}
\end{figure*}

\textcolor{black}{In order to further analyse the effect of age and the effect of interaction order, a two-way ANOVA is conducted to compare the effect of age group and interaction orders on interaction duration across different scenarios. Table \ref{table_example1} indicates that the order of interaction methods does not have a significant effect on the duration of the interaction ($P>0.05$). However, for baseline interaction methods including gesture-based \citep{hanggesture}, NLP-based \citep{stepputtis2020language}, and VLM-based \citep{ConstantinEYBW22}, age groups have a significant effect on interaction duration across scenarios ($P<0.05$). Post-hoc comparisons using the Tukey HSD test \citep{tukey1949comparing} show that there are significant differences in interaction duration between the younger and older groups and almost between all middle and older groups for all interaction methods from the beginning of the study in all scenarios ($P_{adj}<0.05$). In contrast, for our proposed method, age groups have no significant effect on the duration of interaction in different scenarios ($P_{adj}>0.05$). In conclusion, our results suggest that interaction using complex gestures and long sentences takes more time for older users, while our proposed method using only a few common syntaxes is more suitable for older users.}

\begin{figure*}[thp]
\centering
\includegraphics[width=0.8\textwidth]{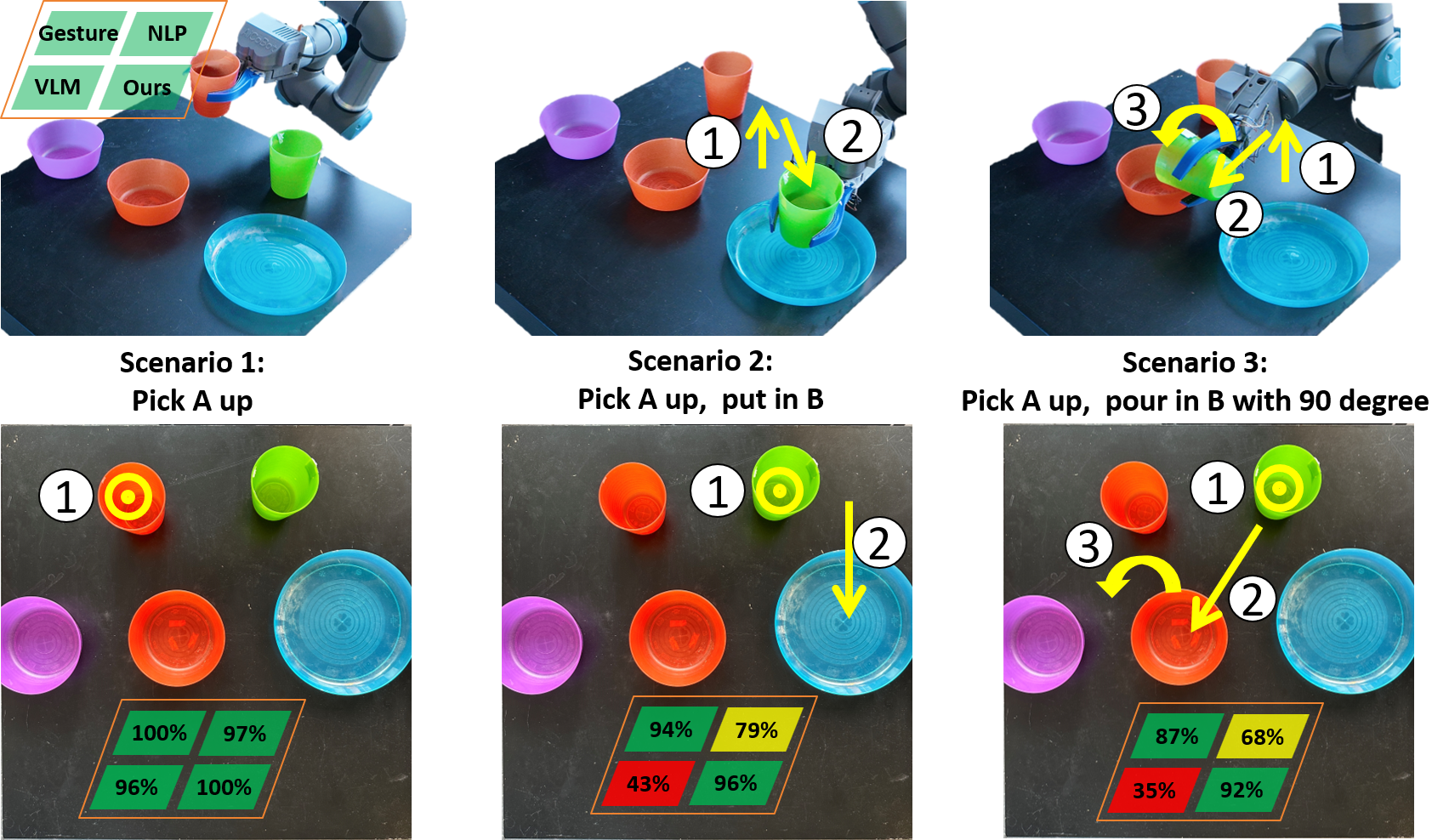}
\caption{\textcolor{black}{Comparison of accuracy of different tasks vs different interaction methods.}} 
\label{fig:acc}
\end{figure*}

\subsection{Comparison of Precision and Robustness}

In this experiment, we placed nine groups of unknown objects, specifically 18 charcoal briquettes arranged in pairs, on a table with 20 cm of separation between them to test HRI accuracy, as illustrated in Figure \ref{fig1:visualfeedback}. Feedback provided by ROS RVIZ visually indicates to the user the specific object being targeted, allowing evaluation of the accuracy of the NVP-HRI interaction in a challenging setting.

The same \textcolor{black}{24} participants from the previous experiment were divided into two groups. All participants received verbal instructions on how to use NVP-HRI. The first group completed the given scenarios three times without visual feedback, resulting in a noticeable improvement in system operation accuracy after brief learning sessions. The second group completed the scenarios with visual feedback, demonstrating that the proposed NVP-HRI could be used effectively without formal training.

\textcolor{black}{To further compare the accuracy of interaction with other baseline methods, we conducted another experiment with participants in different age groups to test the accuracy of different interaction methods. Hundreds of trials were conducted for each interaction method in each scenario during the experiment, and the results are shown in Fig. \ref{fig:acc}. Gesture-based methods perform better than language-based methods because gestures have more tractable reference points. NLP-based and VLM-based methods often require more precise and detailed descriptions and often have challenges in discriminating similar objects. Visual feedback in the proposed method and the gesture-based method also improve the accuracy of the interaction.}
\begin{table}[hp]
\vspace{-8pt}
\caption{\textcolor{black}{Two-Way ANOVA of Interaction Accuracy}}
\label{table_example7}
\begin{center}
\begin{tabular}{cccccccc}
\hline
\hline
\multicolumn{1}{c}{\multirow{3}{*}{Scenario}}&\multicolumn{1}{c}{\multirow{3}{*}{Method}} & \multicolumn{4}{c}{Two-Way ANOVA} \\ \cmidrule(r){3-6}
 && \multicolumn{2}{c}{Age group}  & \multicolumn{2}{c}{Test order} \\ \cmidrule(r){3-4} \cmidrule(r){5-6} 
 & & $F(2,18)$ & $P$  & $F(3,18)$ & $P$ \\ \hline
 \multicolumn{1}{c}{\multirow{4}{*}{S1}}&VLM &0.278&0.760&3.321&0.063 \\ 
&Gesture &0.422&0.661&2.268&0.115 \\ 
&NLP &1.449&0.260&0.308&0.818 \\ 
&\textcolor{black}{NVP-HRI} &0.422&0.661&2.268&0.115 \\ \hline
     \multicolumn{1}{c}{\multirow{4}{*}{S2}}&VLM &2.504&0.109&1.623&0.218 \\ 
      &Gesture &1.348&0.284&0.449&0.720 \\ 
       &NLP &1.226&0.316&0.003&0.999 \\ 
        &\textcolor{black}{NVP-HRI} &1.624&0.224&0.190&0.901\\ \hline
         \multicolumn{1}{c}{\multirow{4}{*}{S3}}&VLM &0.064&0.938&0.298&0.826 \\  
          &Gesture &1.852&0.185&1.848&0.174 \\ 
           &NLP &2.020&0.161&1.922&0.162 \\ 
            &\textcolor{black}{NVP-HRI} &0.403&0.673&0.151&0.927 \\ 
\hline
\hline
\end{tabular}
\end{center}
\end{table}
\textcolor{black}{Table \ref{table_example6} shows the mean interaction accuracy for participants in different age groups to enter the same command in different scenarios with different HRI methods.}

\textcolor{black}{A two-way ANOVA is performed to compare the effect of the age group and the interaction orders on the accuracy of the interaction in different scenarios. Table \ref{table_example7} indicates that the order of interaction methods and the age group do not have a significant effect on  accuracy of interaction ($P>0.05$).}

\begin{table*}[htp]
\caption{\textcolor{black}{HRI Method Preference Survey.}}
\label{table_example}
\begin{center}
\setlength{\tabcolsep}{12pt}
\begin{tabular}{ccccccccccccc}
\hline
\hline
\multicolumn{1}{c}{\multirow{2}{*}{Age groups}} & \multicolumn{3}{c}{NVP-HRI(Ours)} & \multicolumn{3}{c}{NLP}  & \multicolumn{3}{c}{Gesture}  & \multicolumn{3}{c}{VLM}  \\ \cmidrule(r){2-4} \cmidrule(r){5-7} \cmidrule(r){8-10} \cmidrule(r){11-13}
 & $\mathbb{Y}$ & $\mathbb{M}$ & $\mathbb{O}$  & $\mathbb{Y}$ & $\mathbb{M}$ & $\mathbb{O}$  & $\mathbb{Y}$ & $\mathbb{M}$ & $\mathbb{O}$  & $\mathbb{Y}$ & $\mathbb{M}$ & $\mathbb{O}$ \\ \hline
 Comprehensibility &33&15&19&2&2&0&0&0&0&0&1&0 \\ \hline
   Efficiency &114&48&9&0&0&0&0&0&0&0&0&0 \\ \hline
    Simplicity &78&11&23&0&0&0&0&0&0&0&0&0 \\  \hline
    Vogue & 27&3&2&1&0&0&2&0&0&0&0&0\\\hline
    \textbf{Sum} & 252&77&52&3&2&0&2&0&0&0&1&0\\
\hline
\hline
\end{tabular}
\end{center}
\end{table*}
\subsection{Field Survey on User Experience}

To extensively evaluate the proposed solution, a wider community is invited to participate in an online survey\footnote{\url{https://www.wjx.cn/vm/tUHPFD1.aspx}}.
Over 2000 invitations were made, and the invitation mode included email and street surveys conducted with the tablets provided. The candidates comprised professors, nonfaculty staff, and students from around the world. Prior consent was obtained before asking any survey questions. To ensure the accuracy of the delivery of the message, we verified that all candidates were fluent in English. Candidates were asked to indicate their gender, age range and nationality, watch videos demonstrating the same commands being entered using gesture-based, NLP-based, VLM-based, and the proposed HRI method, and then candidates were required to rank the HRI methods based on their preference. Finally, candidates were asked to choose the reasons for their ranking. To minimize bias and order effects, the order of all options was randomized. As a result, only 390 people agreed to complete the survey, which is about a 19.5\% positive response rate. Most of the positive responses are from street inteviews with people in person. At the end of the survey, we invited candidates to experience the proposed interaction framework in a real-world environment and participate in physical experiments. This is done to check the discrepancy between the survey and real-world experience. After participants experienced the real machine HRI, none of the visited participants wanted to change the initial survey result. This indicates that the survey demo video successfully captures the essence of the experiment. 

\begin{figure}[t]
\centering
\includegraphics[width=0.48\textwidth]{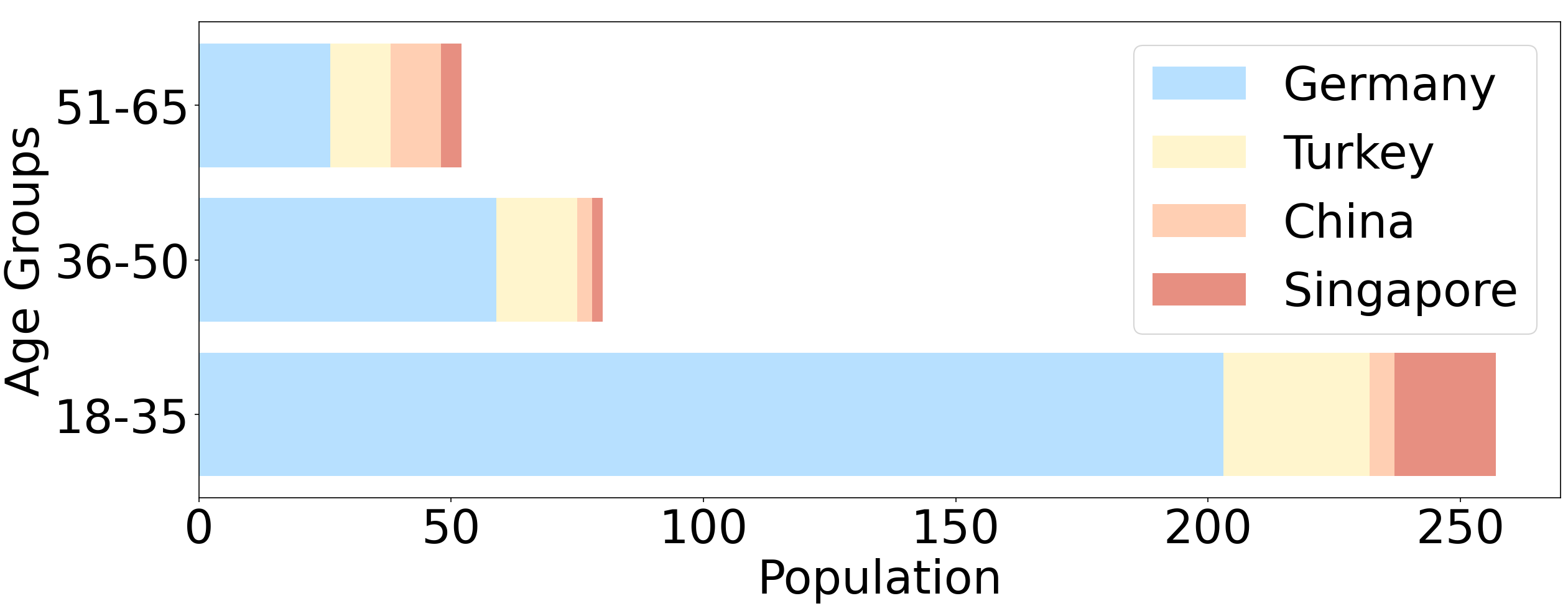}
\vspace{-8pt}
\caption{\footnotesize  Nationality and age distribution of participants.}
\label{fig1:polulation}
\vspace{-15pt}
\end{figure}

To better show the demographic distribution, Fig. \ref{fig1:polulation} shows the nationality and age distribution of the survey participants. The survey results are shown in table \ref{table_example}. The results indicate that more than 97\% of the participants would prefer NVP-HRI over other approaches. Combined with the survey results, participants in the older age groups valued the simplicity and comprehensibility of the interaction methods. Some of the older participants noted that they ranked gesture-based interactions last because it was challenging for them to memorize and perform specific gestures. Some middle-aged and younger participants pointed out the inefficiency of the VLM-based method, which requires identifying an object through dialog with the robot. Participants in the younger and middle-aged age groups valued efficiency more.

\subsection{Diverse Real-world Environment Tests}

\begin{figure}[t]
      \centering
      \includegraphics[width=0.49\textwidth]{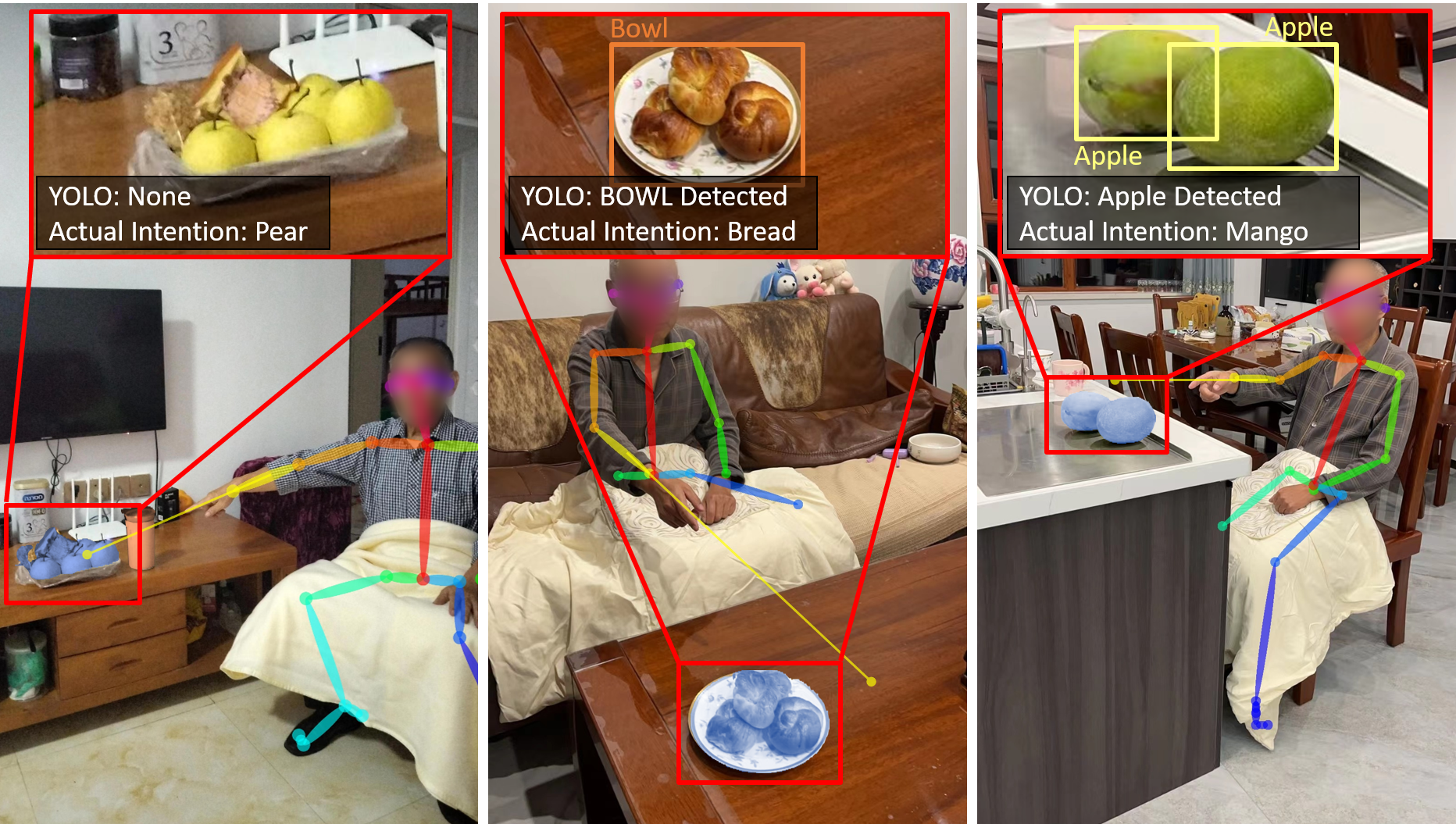}
      \caption{Interaction experiments in the kitchen and living room showed that YOLO consistently failed to detect novel objects accurately, while our proposed method reliably identified these objects through segmentation by SAM and deictic posture estimation via OpenPose.}
      \label{perception}

   \end{figure}

In addition, we conducted HRI experiments in various common scenarios in homes and elderly care centers. In the experiments, we found no difference in the detection of deictic posture between the elderly and the younger participants. Furthermore, when the lower body was obscured, an accurate deictic posture could still be obtained, and HRI accuracy was not affected. In real-world scenarios, supervised models such as YOLO \citep{redmon2016you} often fail to detect or incorrectly categorize novel objects, while our proposed NVP-HRI accurately segments novel objects through SAM \citep{kirillov2023segment}. Our system
successfully retrieved various novel daily objects on demand, as shown in Fig. \ref{perception}.

\section{LIMITATION AND FUTURE WORKS}
Currently, the proposed method has been tested only in laboratory conditions with participants of various age groups. The primary aim of this research is to provide easy access for elderly or ill individuals to control robots to perform daily routines. However, due to the nature of the camera used in this mode, requests to conduct trials in hospitals were denied. Consequently, we have not been able to verify the system's actual performance in real-world settings.

A significant limitation of our study is the demographic profile of our survey participants. The survey was distributed to individuals associated with universities around the world, which inherently biases the sample towards those with higher educational backgrounds and better English fluency. This raises questions about the system's usability among individuals with lower educational levels and different linguistic abilities. It is crucial to conduct further testing in more diverse settings to determine whether the system can be used effectively by a broader population.


Additionally, another challenge is to test the system's effectiveness under low-light conditions. In our experiments, pose recognition is a critical component. However, the performance of pose detection may be compromised in low-light environments. To address this issue, additional testing is necessary. We may also need to integrate supplementary sensors specifically designed to enhance pose detection in low-light conditions.

In summary, while initial laboratory tests of the proposed method are promising, several aspects require further investigation. These include verifying the system's performance in real-world settings, ensuring its usability among a diverse population, and enhancing its effectiveness in low light conditions. Addressing these issues will be crucial to the successful deployment of this technology in practical applications.

\section{CONCLUSIONS}
 
In this work, We introduce NVP-HRI, a multi-modal HRI paradigm combining voice commands and gestures, enabled by the Segment Anything Model (SAM) for scene analysis. NVP-HRI integrates with a large language model (LLM) for real-time trajectory solutions, achieving up to 65.2\% efficiency gains over traditional gesture control. We plan to open-source our code and methodology for greater accessibility.

However, the current system is designed only for English speakers and excludes those with hearing or speaking impairments. Future work will prioritize integrating gestures and additional language models to enhance system versatility.



\printcredits
\section*{Declaration of competing interest}
The authors declare that they have no known competing financial interests or personal relationships that could have appeared to influence the work reported in this paper.

\section*{Acknowledgments}
This work is supported by a grant of the EFRE and MWK ProFö-R\&D program, no. FEIH\_ProT\_2517820 and MWK32-7535-30/10/2. This work is also supported by National Research Foundation, Singapore , under its Medium-Sized Center for Advanced Robotics Technology Innovation.

\section*{Licence and Legal Compliance}

Our code will be released under a Creative Commons Attribution-NonCommercial-ShareAlike 4.0 International License, ensuring its use is limited to non-commercial academic research. The proposed NVP-HRI is shared with the hope that it will significantly benefit the aging community by fostering advancements in human-robot interaction research.

Throughout the project, we have rigorously adhered to the Personal Data Protection Act (PDPA) and the General Data Protection Regulation (GDPR) to protect the privacy and personal data of individuals. To the best of our knowledge, facial masking has been implemented across all data to anonymize the information effectively. We have also appointed Data Protection Officers to oversee and ensure compliance with both PDPA and GDPR standards.

The data is securely hosted on Google Drive, which provides robust security measures to prevent unauthorized access or disclosure. We are committed to maintaining the highest standards of data security and privacy.

Should any concerns arise regarding anonymization, security, or access issues, we strongly encourage individuals to report or request data corrections. Reports or requests can be submitted through the link provided on our GitHub repository, ensuring a transparent and responsive process for addressing any potential issues.

\section*{Data Availability}

Data will be made available upon request. Interested parties can contact us to access the data, ensuring that its use aligns with the stipulated non-commercial and academic research purposes. This process ensures that data sharing is controlled and compliant with our licensing and privacy commitments.
\bibliographystyle{apalike}
\bibliography{cas-refs}




\end{document}